\documentclass[10pt]{iopart}

\usepackage{iopams} 

\usepackage{graphicx} 
\usepackage{threeparttable}
\usepackage{epstopdf}
\usepackage{verbatim}

\begin{document}

\title[TinyGC-Net: An Extremely Tiny Network for Calibrating MEMS Gyroscopes]{TinyGC-Net: An Extremely Tiny Network for Calibrating and Denoising MEMS Gyroscopes}

\author{Cui~Chao$^1$, Jiankang~Zhao$^1$, Long Haihui$^1$, Zhang Ruitong$^1$}

\address{$^1$ 800 Dongchuan RD. Minhang District, Shanghai, China}

\ead{tsuibeyond@sjtu.edu.cn, zhaojiankang@sjtu.edu.cn, longhh@sjtu.edu.cn, zrt19981229@sjtu.edu.cn}
\vspace{10pt}
\begin{indented}
\item[]April 2024
\end{indented}

\begin{abstract}

This paper presents a learning-based method for calibrating and denoising microelectromechanical system (MEMS) gyroscopes, 
which is designed based on a convolutional network, and only contains hundreds of parameters, so the network can be trained on a graphics processing unit (GPU) before being deployed on a microcontroller unit (MCU) with limited computational resources. 
In this method, the neural network model takes only the raw measurements from the gyroscope as input values, and handles the calibration and noise reduction tasks separately to ensure interpretability.
The proposed method is validated on public datasets and real-world experiments, 
without relying on a specific dataset for training in contrast to existing learning-based methods.
The experimental results demonstrate the practicality and effectiveness of the proposed method,
suggesting that this technique is a viable candidate for applications that require IMUs.

\end{abstract}

%
\vspace{2pc}
\maketitle
%
\ioptwocol

\section{Introduction}

Inertial Measurement Units (IMUs) that utilize Microelectromechanical Systems (MEMS) technology have become indispensable in numerous domains, owing to their low power consumption, compact form factor, and cost-effectiveness. These attributes have established MEMS-based IMUs as critical components in various applications, such as robotics \cite{yang2023multi}, mobile devices \cite{wahlstrom2016imu}, wearable technology \cite{majumder2020wearable}, and virtual reality systems \cite{barclay2023characterization}.

Since manufacturers typically implement a rough calibration process for MEMS IMUs to minimize production costs, it becomes imperative for many applications to conduct a more refined IMU calibration before use, which will ensure that the IMU's performance aligns with the precision requirements.

An IMU typically consists of a tri-axial gyroscope for measuring angular velocity and a tri-axial accelerometer for detecting linear acceleration relative to the inertial frame.
For the calibration of tri-axial accelerometers, the local gravitational acceleration serves as a crucial reference, and numerous well-established calibration methods are available.
In contrast, calibrating a tri-axial gyroscope presents significant challenges. This is primarily due to the fact that the Earth's self-rotation angular velocity is weak and often masked by measurement noise. As a result, additional external reference excitation becomes a critical requirement for the calibration of consumer-grade gyroscopes. 

Over recent years, gyroscope calibration has garnered widespread interest and undergone thorough investigation within the open-source community. 
Many researchers have devoted their efforts to developing calibration methods that are both efficient and accurate, aiming to improve the measurement precision of gyroscopes.
In general, gyroscope calibration techniques can be categorized into two primary groups: classical and learning-based methods. 

Classical methods are distinguished by their use of explicit and rigorous mathematical models to represent the sensor's measurements.
For instance, Zhang et al. introduced a thorough sensor model for inertial devices, where gyroscope calibration is facilitated through a pan-tilt mechanism \cite{zhang2013calibration}; 
Qureshi et al. presented a novel calibration algorithm tailored for IMUs, which operates independently of external equipment and is suitable for use outside controlled laboratory settings \cite{qureshi2017algorithm}; 
Chao et al. devised a calibration approach that requires minimal setup for low-cost tri-axial IMUs and magnetometers, and the tri-axial gyroscope is calibrated by utilizing a pre-calibrated accelerometer and a nonlinear cost function \cite{chao2021minimum}. 
Furthermore, system-level calibration methodologies have been extensively implemented and validated, albeit with a reliance on carefully designed rotation sequences. 
For instance, ghanipoor et al. utilizes the Transformed Unscented Kalman Filter (TUKF) in conjunction with a turntable setup to perform calibration on MEMS IMUs \cite{ghanipoor2020toward}; 
Lu et al. introduced a comprehensive system-level calibration technique for Stellar/Inertial Navigation Systems within the Kalman filter framework, and incorporates a 12-position rotation scheme to guarantee that all system parameters are adequately observable, thereby enhancing the calibration accuracy \cite{lu2017all}; 
Jung et al. used the extended Kalman filter (EKF) to analyze the observability of IMU error parameters in a stereo visual-inertial system, and consider that the IMU's intrinsic parameters are observable when the system undergoes 6 degrees of freedom motions \cite{jung2020observability}. 

Essentially, classical methods tend to treat the sensor's measurement model as inherently linear , which encounters considerable difficulty when attempting to adapt to nonlinear frameworks.
To overcome this limitation, learning-based approaches have been introduced for the calibration of MEMS IMUs, which have attracted considerable attention from the research community as an innovative solution. 
So far, most studies have attempted to utilize deep neural networks to extract relevant features from IMU readings to regress the navigation results, such as IONet \cite{chen2018ionet}, OriNet \cite{esfahani2019orinet}, RoNIN \cite{herath2020ronin}, TLIO \cite{liu2020tlio}. 
And only a few researchers try to use the deep learning technology to calibrate IMUs directly \cite{brossard2020denoising, huang2022mems},
who have devised an elaborate calibration network for MEMS IMUs, and have indicated that it is feasible to use deep learning technology to eliminate IMU errors. 

Despite notable calibration improvements demonstrated by existing learning-based methods, the trained networks with tens of thousands of parameters can only be operated on graphics processing units (GPUs) or desktop-grade central processing units (CPUs), thereby significantly limiting the scope of their applications.
On the other hand, all learning-based methods require the accelerometer's measurements as inputs to the network to restrain the gyroscope's time-varying bias, which may lead to anomalous outputs under the extreme dynamic conditions. For instance, the accelerometer's measurement will approach zero in the case of free-fall.
Furthermore, the existing learning-based methods heavily rely on open-source datasets for training, thereby limiting their practical application in industrial scenarios due to the significant challenges and high costs associated with obtaining reference attitude angles.

Therefore, we propose an ultra-lightweight network for calibrating and denoising gyroscopes, which is referred to as TinyGC-Net in this paper.
Given that the contribution of the denoising is nearly overshadowed by the influence of the integral step in typical navigation scenarios \cite{ban2013integral}, TinyGC-Net reduces the model's parameter count by handling gyroscope calibration and denoising tasks separately.
Furthermore, TinyGC-Net demonstrates effectiveness in modeling the measurement process of gyroscopes, and is the first learning-based solution for gyroscope calibration suitable for deployment on resource-limited embedded platforms.

The main contributions of this work are as follows:

(1) We propose a learning-based method for calibrating and denoising gyroscopes, and the trained model is ultra lightweight, containing only hundreds of parameters,  which can be deployed on a low-cost processor with limited computing resources and run in real-time.

(2) The cost function is carefully designed to empower TinyGC-Net to operate independently of open-source datasets, thereby facilitating the efficient collection of training data solely through the use of a tri-axial manual rotation table or a pre-calibrated accelerometer.

(3) The proposed method demonstrates outstanding interpretability, enabling its reliable operation in various industrial application scenarios.

The rest of this paper is organized as follows: 
Section \ref{Sec_Preliminaries} describes the measurement model of the gyroscopes. 
Section \ref{Sec_Proposed_Method} provides a detailed exposition of the TinyGC-Net. 
Section \ref{Sec_Experiments} shows the experimental results based on the public datasets and the data from the designed calibration process.
In Section \ref{Sec_Conclusion}, we draw the concluding remarks.

\section{Preliminaries}\label{Sec_Preliminaries}

\subsection{Gyroscope Measurement Model}

A classical linear equation can describe the relationship between the gyroscope voltage readings sampled by an analog-to-digital converter (ADC) and the physical quantities in metric units. It can be written as follows:

\begin{equation}\label{eqA1}
	\tilde{\boldsymbol{\omega}}=\boldsymbol{R}_g \boldsymbol{T}_g \boldsymbol{S}_g\left(\boldsymbol{\omega} + \boldsymbol{b}_g + \boldsymbol{n}_g\right),
\end{equation}

\noindent where the subscript $g$ indicates the sensor's type is the gyroscope;
$\tilde{\boldsymbol{\omega}}\in \mathbb{R}^{3\times 1}$ is the measured angular velocity in the unit of $rad/s$;
$\boldsymbol{\omega}\in \mathbb{R}^{3\times 1}$ is the sensor voltage readings sampled by ADC;
$\boldsymbol{R}_g$,  $\boldsymbol{T}_g$ and $\boldsymbol{S}_g$ are $3\times 3$ matrices,  which account for misalignment, non-orthogonality, and scale factor, respectively; 
$\boldsymbol{b}_g \in \mathbb{R}^{3\times 1}$  is the bias vector; 
$\boldsymbol{n}_g\in \mathbb{R}^{3\times 1}$ is zero-mean Gaussian noises. 
As the Earth's angular rate is too small to be detected by a MEMS gyroscope and is often overwhelmed by white noise, equation (\ref{eqA1}) does not consider it. 

Further, equation (\ref{eqA1}) can be simplified as follows

\begin{equation}\label{eqA2}
	\tilde{\boldsymbol{\omega}}=\boldsymbol{E}_g \boldsymbol{\omega}+\boldsymbol{B}_g + \boldsymbol{\eta}_g,
\end{equation}

\noindent where 

$$
\boldsymbol{E}_g = \boldsymbol{R}_g\boldsymbol{T}_g\boldsymbol{S}_g=\left(\begin{array}{lll}
	e_{00} & e_{01} & e_{02} \\
	e_{10} & e_{11} & e_{12} \\
	e_{20} & e_{21} & e_{22}
\end{array}\right),
$$

$$
\boldsymbol{B}_g = \boldsymbol{R}_g\boldsymbol{T}_g\boldsymbol{S}_g \boldsymbol{b}_g = \left(\begin{array}{l}
	b_0 \\
	b_1 \\
	b_2
\end{array}\right),
$$

\noindent $\boldsymbol{\eta}_g \in \mathbb{R}^{3\times 1}$ is still assumed to be zero-mean Gaussian noise. 
And the essence of traditional gyroscope calibration is to estimate the elements in $\boldsymbol{E}_g$ and $\boldsymbol{B}_g$.

In the actual gyroscope measurement model, $\boldsymbol{S}_g$ exhibits a certain degree of nonlinearity.
Hence, the elements within matrices $\boldsymbol{E}_g$ and $\boldsymbol{B}_g$ can also be represented by higher-order polynomials \cite{ru2022mems}. However, traditional calibration methods struggle to effectively estimate all the polynomial coefficients.
In light of this, this paper employs a Convolutional Neural Network (CNN) to model the measurement model of the gyroscope.

\subsection{Kinematic Model}

We choose the north-east-down frame as the navigation coordinate frame ($n$), whose $x_n$-axis points to the geodetic north, $y_n$-axis points to the geodetic east, and $z_n$-axis points downwards. And the body coordinate frame ($b$) is fixed to the center of the IMU sensor, with its $x_b$-axis, $y_b$-axis, and $z_b$-axis pointing in the forward, right, and downward directions, respectively.

The 3D orientation of a rigid platform is obtained by integrating the gyroscope's angular velocity measurements, and we use quaternions to describe the kinematic model as follows:

\begin{equation}\label{eqA3}
	\mathbf{q}_{k+1}=\mathbf{q}_k \otimes\left[\begin{array}{c}
		1 \\
		\frac{1}{2} \tilde{\boldsymbol{\omega}}_k dt
	\end{array}\right],
\end{equation}

\noindent where the quaternion $\mathbf{q}_{k}$ at timestamp $k$ represents the rotation from the body coordinate frame to the navigation coordinate frame; 
$\tilde{\boldsymbol{\omega}}_k$ is the mean angular velocity during the time period $dt$ from $k$ to $k+1$. 
The model (\ref{eqA3}) successively integrates in open-loop and propagates the estimation errors caused by the inaccurate $\boldsymbol{E}_g$ and $\boldsymbol{B}_g$. 

\section{Proposed Method}\label{Sec_Proposed_Method}

In this section, we propose a learning-based method for obtaining more accurate angular velocity measurements. 
Compared with existing learning-based methods, TinyGC-Net aims to reduce the number of parameters in the network model and simplifies the process by dividing it into: a calibration subnet and a denoising subnet. 

\begin{figure}[htbp]  
	\centering
	\includegraphics[width=7.5cm]{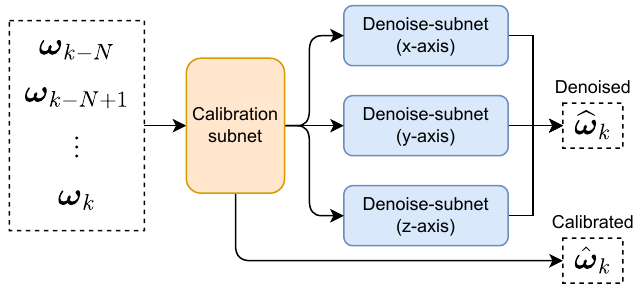}
	\caption{The overall network architecture of TinyGC-Net.}\label{fig1}
\end{figure}

The overall network architecture is depicted in Figure \ref{fig1}, 
where $\boldsymbol{\omega}_{k}$ is the voltage reading from the tri-axis gyroscope at timestamp $k$; 
$N$ is the length of the local window;  
$\hat{\boldsymbol{\omega}}_{k}$ and $\widehat{\boldsymbol{\omega}}_{k}$ are the calibrated and denoised angular velocity, respectively, in the unit of $rad/s$ at timestamp $k$.

\subsection{Calibration Subnet}

\begin{figure}[htbp]  
	\centering
	\includegraphics[width=5.5cm]{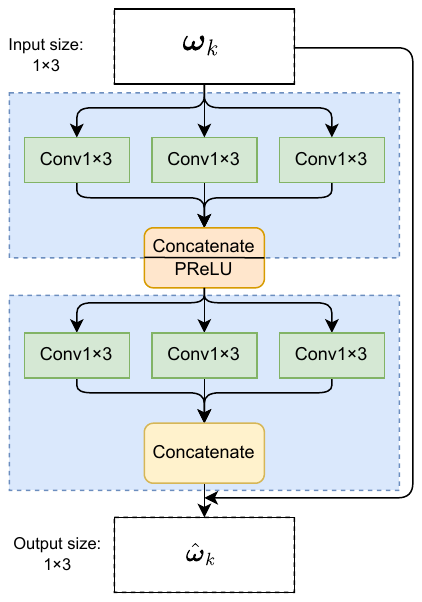}
	\caption{The network architecture of calibration subnet.}\label{fig2}
\end{figure}

The calibration subnet is constructed as in Figure \ref{fig2}. 
Essentially, the convolutional and concatenation operations within the dashed box are equivalent to equation (\ref{eqA1}) if all the parameters are constant, and we refer to this as the linear basic network (LBN).

To accurately model the nonlinear components of the gyroscope measurement model, we employ the PReLU activation function \cite{ding2018activation} to connect LBN modules, drawing inspiration from the structure of residual networks to facilitate faster convergence during training.

Due to the continuous advancements in MEMS process technology, most manufacturers' MEMS gyroscopes exhibit nonlinearity within a range of 0.01\% in terms of angular velocity measurement. 
On the other hand, according to our experimental observations, the cascading of multiple LBNs can hardly enhance calibration accuracy. Therefore, in the subsequent experiments detailed below, we only incorporate two LBNs into TinyGC-Net.

\subsection{Denoising Subnet}

Through approximation, we assume that the tri-axial measurement processes of the gyroscope are independent of each other, allowing us to individually apply noise reduction processing to the measurement data of each axis. 

Then, we define the denoising subnet structure which infers the denoised gyroscope's measurements as

\begin{equation}
	\widehat{\omega}_{i, k}=f\left(\hat{\omega}_{i, k-N+1}, \cdots, \hat{\omega}_{i, k}\right),
\end{equation}

\noindent where the subscript $i$ indicates the $i$-axis of the gyroscope ($i=x, y, z$);
$N$ is the length of the local window; 
$\hat{\omega}_{i, k}$ is the calibrated angular velocity of the gyroscope's $i$-axis at timestamp $k$; 
$f(\cdot)$ is the mapping defined by the neural network; 
$\widehat{{\omega}}_{i, k}$ is the denoised data. 

\begin{figure}[htbp]  
	\centering
	\includegraphics[width=7.5cm]{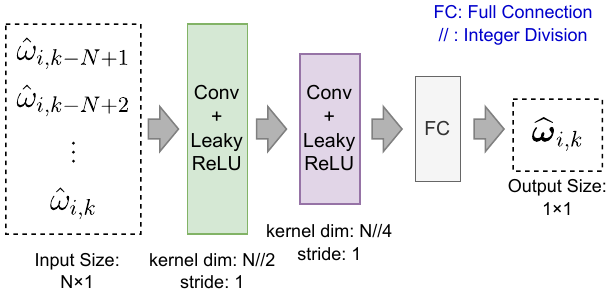}
	\caption{The network architecture of denoising subnet.}\label{fig3}
\end{figure}

The denoising subnet consists of three convolutional layers and three LeakyReLU layers, whose configurations are given in Figure \ref{fig3}. 
Essentially, we leverage the denoising subnet that infer denoised data based on a local window of $N$ previous measurements. 
Since there are no coupled parameters among the three axes of the gyroscope, the denoising subnet requires only hundreds of parameters in total to denoise the measurements for each axis individually.

\subsection{Loss Function}

\begin{figure}[htbp]  
	\centering
	\includegraphics[width=7.5cm]{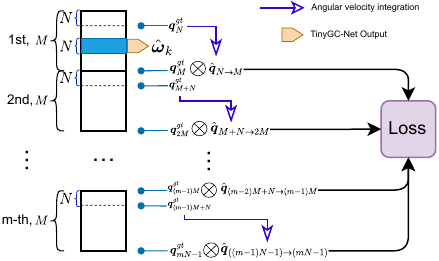}
	\caption{The architecture of loss function.}\label{fig4}
\end{figure}

Defining a proper loss function plays a critical role in training the calibration parameters within the neural network.
Given that it is difficult and costly to obtain the ground truth values for every moment, 
we construct a simple loss function shown in Figure \ref{fig4}, which only requires a few pieces of reference attitude information over a period of sampling time, and it is also convenient for real-world scenarios without a motion capture system.

In Figure \ref{fig4}, the quaternion $\boldsymbol{q}_{N}^{gt}$ represents the ground truth of the attitude angle at timestamp $N$;
$\hat{\boldsymbol{q}}_{N\to M}$ represents the estimated attitude angle that is based on the equation (\ref{eqA3}) from timestamp $N$ to $M$; 
the symbol $\ominus$ represents the difference between two quaternions, and we define it as follows in this paper:

$$
\boldsymbol{q_a} \ominus \boldsymbol{q_b} = \sqrt{d{q}_{w}^2 + d{q}_{x}^2 + d{q}_{y}^2 + d{q}_{z}^2},
$$

\noindent where $\boldsymbol{q}_a = \left(q_{a,w}, q_{a,x}, q_{a,y}, q_{a,z}\right)$ and $\boldsymbol{q}_b = \left(q_{b,w}, q_{b,x}, q_{b,y}, q_{b,z}\right)$ are two distinct quaternions;
$q_{a,w}$ and $q_{b,w}$ are the scalar parts;
$\left(q_{a,x}, q_{a,y}, q_{a,z}\right)$ and $\left(q_{b,x}, q_{b,y}, q_{b,z}\right)$ are the vector parts;
and
$$
dq_s=q_{a, s}-q_{b, s},\quad s=w, x, y, z.
$$

As illustrated in Figure \ref{fig4}, the sequences of gyroscope measurements, each with a length of $mM$, will be segmented into $m$ equal parts, each part being $M$ in length. Within each segment, TinyGC-Net employs a local sliding window mechanism to analyze the raw gyroscope measurement data within the local window of length $N$, and performs local noise reduction and calibration. 
This localized processing aids in capturing short-term dynamic changes in the gyroscope measurement data and suppresses high-frequency noise within the measurements. 

For the $j$-th gyroscope measurement sequence of length $M$, the cost function can be represented as follows:

\begin{equation}
	\mathcal{L}_{j} = \boldsymbol{q}_{jM}^{gt} \ominus \hat{\boldsymbol{q}}_{(j-1)M+N \to {jM}},
\end{equation}

During the training process, the calibration subnet and the denoising subnet are trained separately, but they utilize a shared cost function. Initially, we temporarily set $N$ to 1 and focus exclusively on training the calibration subnet. Afterward, we fix the weights of the calibration subnet and proceed to train the denoising subnet.

\subsection{Method Implementation}

The network of this paper is trained using PyTorch on a desktop computer with a i7-13790F CPU and a Nvidia GeForce RTX 4060 Ti GPU. The training process uses the AdamW optimizer \cite{loshchilov2017decoupled}, and set the learning rate to 0.01.

In the training process, the whole network only requires hundreds of parameters to be trained, which is significantly fewer than other calibration models based on deep learning. 
And the 2000 epochs of training take about 15 minutes ($N = 50$, $M=400$).


\section{Experiments}\label{Sec_Experiments} 

To verify the validity of the TGC-Net, we carry out a series of experiments based on public datasets, as well as real-world scenarios, in this section.

\subsection{Experiments on public datasets}

\subsubsection{Data Sources and Training Details}

The European Robotics Challenge (EuRoC) micro air vehicle (MAV) dataset \cite{burri2016euroc} is a visual-inertial dataset, which contains IMU sequences at 200Hz from an ADIS16448 MEMS IMU sensor, and an external Vicon system is used to provide the pose ground truth.
For ease of comparison with previous work, we use MH\{01, 03, 05\}, V1\{02\}, and V2\{01, 03\} from the EuRoC dataset for training, while the rest are used for testing.

\subsubsection{Metrics Definitions}

We utilize absolute orientation error (AOE) \cite{grupp2017evo} to quantitatively assess the performance of the proposed method.
The AOE computes the mean square error between the ground truth and the estimated orientation, which can be described as follows: 

\begin{equation}
\mathrm{AOE}=\sqrt{\sum_{n=1}^L \frac{1}{L}\left\|\log \left(\mathbf{R}_n^T \hat{\mathbf{R}}_n\right)\right\|_2^2},
\end{equation}

\noindent where $\mathbf{R}_n\in SO(3)$ is the rotation matrix at timestamp $n$ maps the body coordinate frame to the navigation coordinate frame, and $\hat{\mathbf{R}}_n\in SO(3)$ is its estimated value; 
$L$ represents the sequence length; 
$\log(\cdot)$ is the $SO(3)$ logarithm map.

\subsubsection{Compared Methods}

We compare the following methods based on the dataset:

(1) \textbf{Raw}: the attitude angles are calculated using raw gyroscope measurements without any correction or calibration applied;

(2) \textbf{DIG} \cite{brossard2020denoising}: the IMU denoising method based on a dilated convolutional neural network;

(3) \textbf{OriNet} \cite{esfahani2019orinet}: the 3D orientation estimation method based on long short-term memory (LSTM), which is more complex and requires more time to be trained than DIG;

(4) \textbf{TinyGC}: our proposed method described in Section \ref{Sec_Proposed_Method}. 


\subsubsection{Experimental results}

\begin{table*}
	\centering
	\caption{Absolute Orientation Error (AOE) in degree on the test sequences.}\label{table1}
	\scriptsize
	\begin{threeparttable}
		\begin{tabular}{cccccccc} 
			\hline
			& Raw  & DIG \cite{brossard2020denoising}  & OriNet \cite{esfahani2019orinet} & \begin{tabular}[c]{@{}c@{}}TinyGC-Net\\(Calibrated)\end{tabular} & \begin{tabular}[c]{@{}c@{}}TinyGC-Net\\(Denoised)\end{tabular} \\ 
			\hline
			MH\_02\_easy      & 146  & \textbf{1.39} & 5.75   & 6.36   & 4.60     \\
			MH\_04\_difficult & 130  & \textbf{1.40}  & 8.85   & 5.32   & 4.16    \\
			V1\_01\_easy      & 71.3 & \textbf{1.13} & 6.36   & 7.12  & 6.27     \\
			V1\_03\_difficult & 119  & 2.70  & 14.70   & 3.64   & \textbf{2.23}  \\
			V2\_02\_medium    & 117  & \textbf{3.85} & 11.70   & 4.95   & 5.03  \\
			parameter count   & -    & 77052 & -     & \textbf{27}    & 195   \\
			model input & - &  ACC \& GYRO & ACC \& GYRO & \textbf{GYRO} & \textbf{GYRO}  \\
			training platform & -    & GPU  & GPU    & GPU    & GPU  \\
			deployment platform & -  & GPU  & GPU    & \textbf{MCU} & \textbf{MCU} \\
			\hline
		\end{tabular}
		
		\begin{tablenotes}
			\footnotesize
			\item[*] OriNet uses LSTM network, and its  model is more complex and require more time to be trained than DIG.
		\end{tablenotes}
		
	\end{threeparttable}
\end{table*}

\begin{figure}[htbp]  
	\centering
	\includegraphics[width=7.5cm]{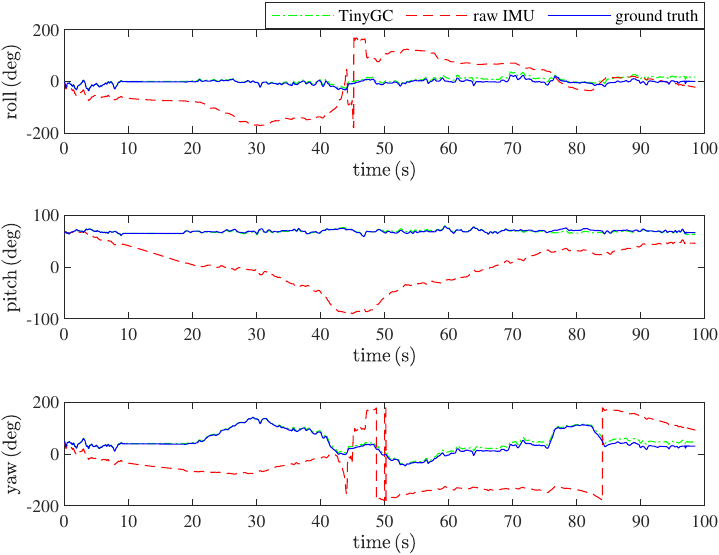}
	\caption{Orientation estimations on the test sequence V2\_02\_medium.}\label{fig5}
\end{figure}

\begin{figure}[htbp] 
	\centering
	\includegraphics[width=7.5cm]{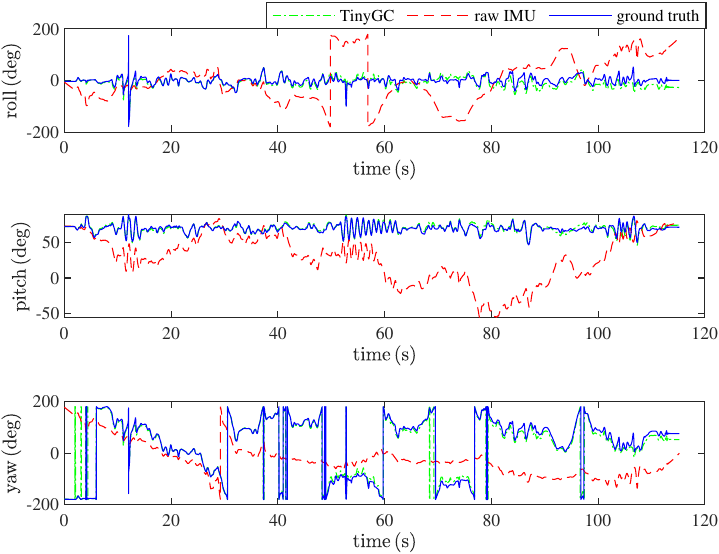}
	\caption{Orientation estimations on the test sequence MH\_04\_difficult.}\label{fig6}
\end{figure}

\begin{figure}[htbp]  
	\centering
	\includegraphics[width=7.5cm]{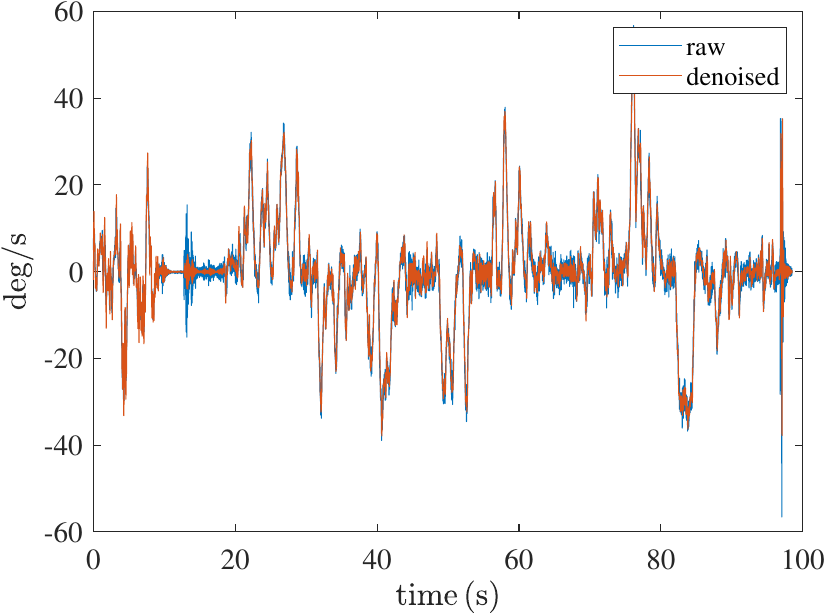}
	\caption{Comparison of time-domain sequences of gyroscope measurements before and after denoising on the test sequence MH\_04\_difficult.}\label{fig7}
\end{figure}

\begin{figure}[htbp]  
	\centering
	\includegraphics[width=7.5cm]{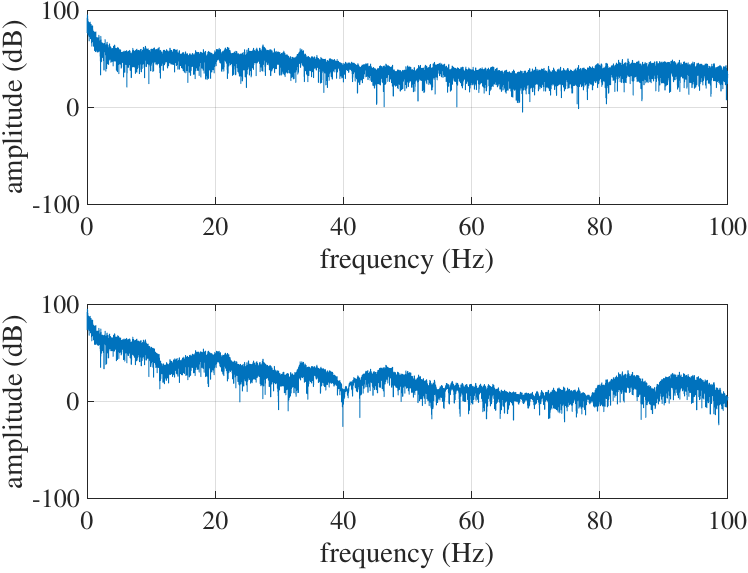}
	\caption{Comparison of gyroscope measurements in the frequency domain before and after denoising on the test sequence MH\_04\_difficult.}\label{fig8}
\end{figure}

The experimental results are given in Table \ref{table1}. 
In contrast to other methods, TinyGC-Net uniquely depends on gyroscope measurements to calculate attitude angles, and offers improved efficiency and reduced computational burden. 
Despite its accuracy trailing DIG, TinyGC-Net outperforms OriNet in precision across most of the sequences. 
Figures \ref{fig5} and Figure \ref{fig6} clearly depict the attitude angles derived from integrating gyroscope measurements calibrated by TinyGC-Net, which are based on the sequences V2\_02\_medium and MH\_04\_difficult.
Furthermore, we also demonstrate the denoising effect on the gyroscope measurement data MH\_04\_difficult in Figure \ref{fig7},
and the spectral analysis results are presented in Figure \ref{fig8},
revealing that the high-frequency noise in the original gyroscope measurements has been effectively suppressed.

According to the above experimental results, we note that:

(1) TinyGC-Net demands minimal parameters and is the only scheme among all the solutions that can be implemented on MCUs with limited computational resources;

(2) DIG often achieves the highest accuracy but requires a GPU for both training and operation;

(3) Overall, OriNet does not exhibit a significant accuracy advantage compared to TinyGC-Net, despite TinyGC-Net solely relying on gyroscope measurements as model input;

(4) The uncalibrated gyroscope measurements are unreliable, and the integration of raw data to derive orientation angles is susceptible to rapid drifting;

(5) The denoising effect is clearly observed, effectively mitigating high-frequency noise in the gyroscope measurements.

\subsubsection{Remark}

We provide a few more remarks based on the process of our research: 

(1) Essentially, most learning-based gyroscope calibration methods rely on accelerometer data and sliding window measurements to constrain offsets and improve orientation accuracy. 
However, this approach will increase model complexity, and demand regularization techniques like weight decay and dropout to prevent overfitting.
Additionally, the data samples employed for network training cannot comprehensively cover all movement patterns, particularly instances like free fall where accelerometer readings tend towards zero, which may elevate the chances of abnormal model outputs. 

(2) Two LBNs are adequate for most MEMS gyroscope calibration tasks. Incorporating additional LBNs and activation functions into the calibration sub-network yielded minimal improvement in attitude angle estimation accuracy. Furthermore, the ADIS16488, a tactical-grade inertial sensor used in the EuRoC dataset, exhibits gyroscope nonlinearity of only $0.01\%$ of the dynamic range, as specified in its datasheet.

(3) Compared to the classic low-pass filter, TinyGC-Net requires minimal parameter tuning and can automatically balance noise reduction and delay based on training samples. 

(4) Drawing from the experimental results presented in Table \ref{table1}, we exercise caution in highlighting that TinyGC-Net's denoising subnet has the potential to enhance the attitude angle estimation accuracy, while the significance of denoising is largely overshadowed by the impact of the integral step \cite{ban2013integral}.
In our opinion, the EuRoC datasets, primarily designed for assessing visual odometry algorithms, may exhibit limitations in IMU device calibration. 
According to the known issues provided by \cite{burri2016euroc}: ``some of the datasets exhibit very dynamic motions, which are known to deteriorate the measurement accuracy of the laser tracking device.  
And the numbers reported by the manufacturer maybe overly optimistic for these events, which complicated the interpretation of ground truth comparisons for highly accurate visual odometry approaches."
Therefore, the improved accuracy in attitude angle estimation can thus be attributed to the sliding window mechanism of the denoising subnet, which effectively mitigates, to a certain extent, errors resulting from dataset imperfections.

\subsection{Experiments on real-world scenarios}

It is worth mentioning that deep learning-based gyroscope calibration algorithms depend heavily on open-source datasets for reference orientation angles, limiting their applicability in diverse scenarios. 
To address this limitation, we introduce a straightforward calibration steps that employs a manually controlled tri-axial turntable. 

\subsubsection{Experiment setup}

\begin{figure}[htbp]  
	\centering
	\includegraphics[width=7.5cm]{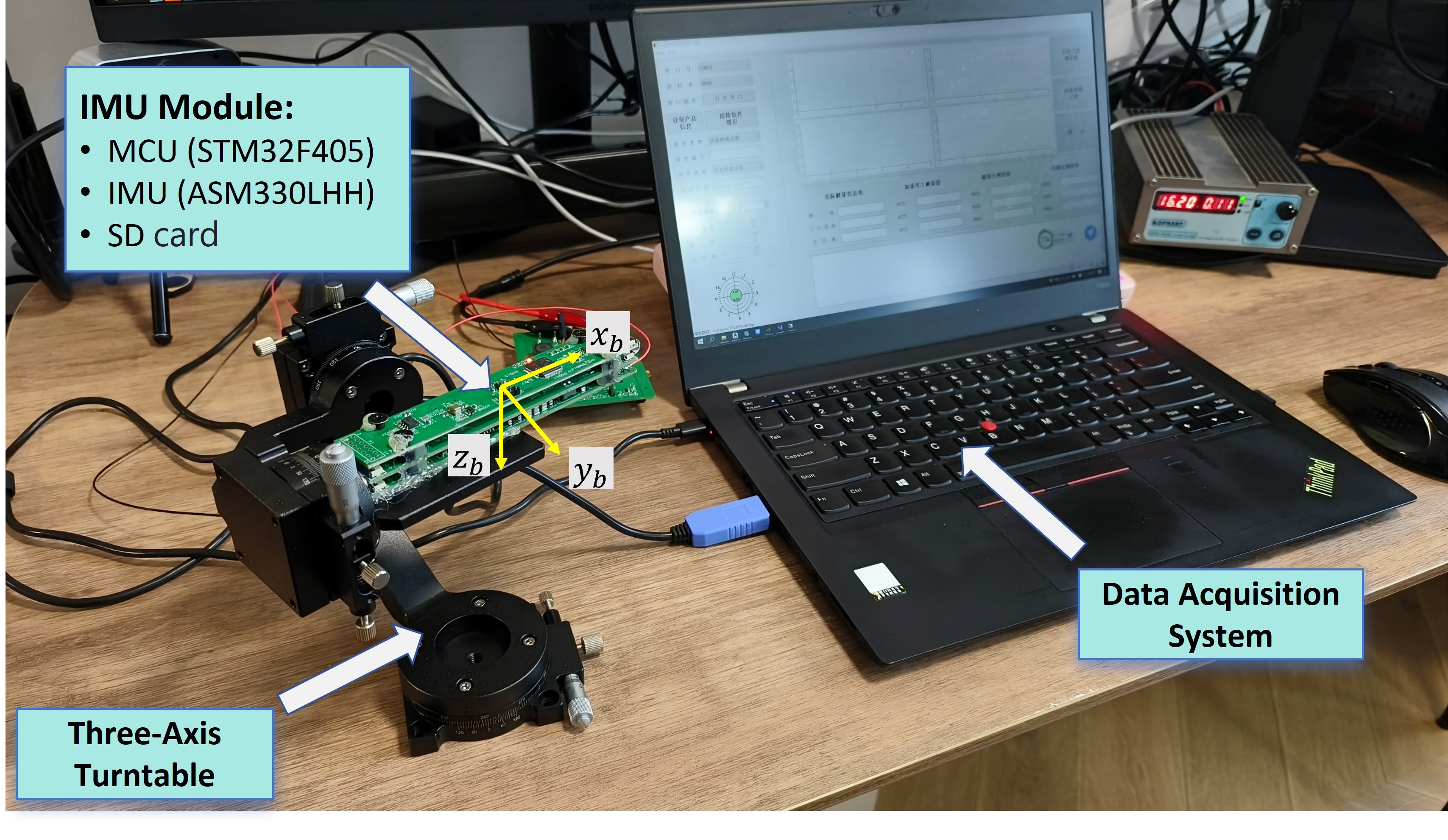}
	\caption{The custom-made experiment system.}\label{fig9}
\end{figure}

A custom-made experiment system is built to collect measurements from the IMU and assess the performance of TinyGC-Net.
As shown in Figure \ref{fig9}, the experiment system is equipped with an IMU (ASM330LHH) for acceleration and angular velocity measurements, whose main specifications are summarized in Table \ref{Table2}.
During the experiments, original measurements are read by a MCU (STM32F405RGT6)
and recorded on the module's secure digital (SD) card.

\begin{table}
	\centering
	\caption{Specifications of the MEMS gyroscope (ASM330LHH)}\label{Table2}
	\small
	\begin{tabular}{ccc} 
		\hline
		Parameter              & Value & Conditions  \\ 
		\hline
		Full scale range      & ±500$^\circ/s$     &            \\
		Nonlinearity           & ±0.01\%     &   25$^\circ $C         \\
		Sensitivity tolerance  & ±5\%     & Component level           \\
		Bias instability & 3 deg/h & 25$^\circ $C \\
		Rate noise density & 5 mdps/$\sqrt{Hz}$ &  \\
		\hline
	\end{tabular}
\end{table}

\begin{figure}[htbp] 
	\centering
	\includegraphics[width=7.5cm]{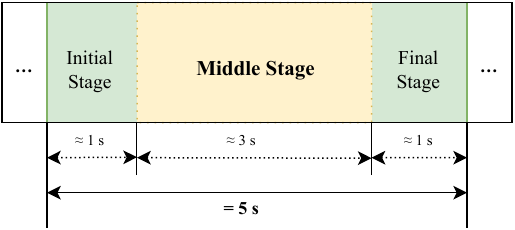}
	\caption{The data collection mode for training TinyGC-Net.}\label{fig10}
\end{figure}

Before collecting gyroscope measurement data, the IMU is securely mounted on the tri-axial turntable. 
Subsequently, data collection for training purposes is divided into three phases: the initial, intermediate, and final phases, as depicted in Figure \ref{fig10}.

In the initial and final phases, the IMU remains stationary for approximately one second to obtain a dependable reference orientation angle.
Meanwhile, in the intermediate phase, the tri-axial turntable undergoes approximately three seconds of manual random rotation.
Each sequence segment is consistently set to 5 seconds to facilitate GPU-based training.

Through the repeated execution of the initial, intermediate, and final phases, we effectively collect 40 data segments to train the TinyGC-Net model, which can subsequently be deployed and operationalized on the MCU.

Afterwards, to assess the performance of TinyGC-Net based on a real-world scenario, we utilize the turntable to gather supplementary IMU measurements spanning 58 seconds, with the guarantee that the turntable's initial and final attitude angles are both set at zero degrees. 

It is worth mentioning that the process mentioned earlier only demands the reference attitude angles of the initial and final stages. Consequently, in practical applications, relying exclusively on a pre-calibrated accelerometer to provide these reference attitude angles is feasible.

Furthermore, we also utilize the Particle Swarm Optimization (PSO) algorithm to calibrate the gyroscope \cite{chao2021minimum}, 
which is renowned for its capability to converge towards theoretical true values through extensive computational processes, thereby yielding refined calibration models for gyroscopes.

\begin{figure}[htbp]  
	\centering
	\includegraphics[width=7.5cm]{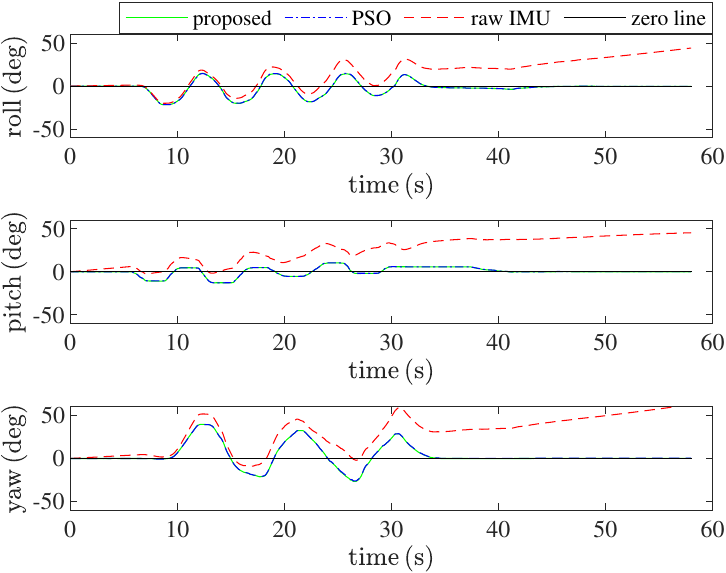}
	\caption{Orientation estimations on real-world scenario.}\label{fig11}
\end{figure}

According to Figure \ref{fig11} and Table \ref{Table3}, it is evident that both TinyGC-Net and PSO adeptly calibrate the gyroscope, with TinyGC-Net exhibiting superior precision.
The remarkable consistency in calibration accuracy between TinyGC-Net and the PSO algorithm highlights the efficacy of deep learning in refining gyroscope measurement models, illustrating the reliability of the TinyGC-Net in real-world scenarios. 
And the Z-axis error is slightly larger than that of the X and Y axes, potentially due to manufacturing imperfections in low-end MEMS IMUs or uneven welding stress on the printed circuit board.

\begin{figure}[htbp]  
	\centering
	\includegraphics[width=7.5cm]{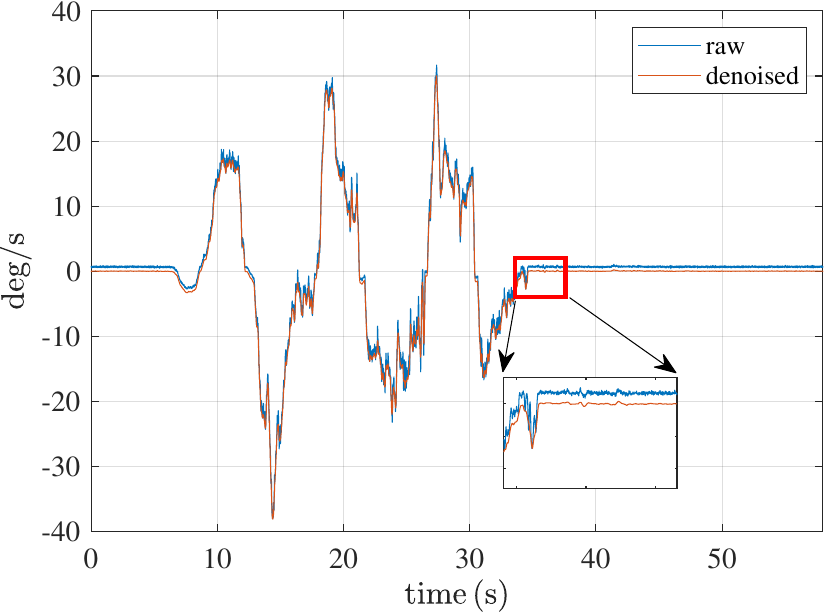}
	\caption{Comparison of gyroscope measurements before and after denoising on real-world scenario.}\label{fig12}
\end{figure}

Figure \ref{fig12} clearly exhibits the noise reduction achieved by TinyGC-Net in gyroscope measurements, successfully maintaining genuine angular motion data while mitigating high-frequency noise, which holds positive implications for applications such as control systems and the stationary alignment of INS.

\begin{table}
	\centering
	\caption{\centering{Orientation error of the final attitude angles compared with zero degree.}}\label{Table3}
	\begin{tabular}{cccc}
		\hline
		& raw   & PSO     & TinyGC-Net \\ 
		\hline
		Roll(deg)    & $62.68^\circ$ & $0.29^\circ$  & $\textbf{0.04}^\circ$   \\
		Pitch(deg)   & $45.34^\circ$ & $0.23^\circ$  & $\textbf{0.06}^\circ$   \\
		Yaw(deg)     & $44.27^\circ$ & $-0.29^\circ$ & $\textbf{-0.25}^\circ$  \\
		RMSE(deg)    & $51.45^\circ$ & $0.27^\circ$  & $\textbf{0.15}^\circ$   \\
		\hline
	\end{tabular}
\end{table}

\section{Conclusion}\label{Sec_Conclusion}

In conclusion, we have presented TinyGC-Net, which is based on deep learning for calibrating and denoising tri-axis gyroscopes. It only requires a few hundred parameters in total and can be deployed on an MCU with limited computational resources.
The core of this approach lies in the careful design of a convolutional network to handle the tasks of calibration and noise reduction separately,
and an appropriate loss function design for training with orientation reference at extremely low frequency.
Compared with other calibration methods based on machine learning, the proposed TinyGC-Net does not rely on the accelerometer's measurement and does not require a specific dataset for training.
So, TinyGC-Net is adaptable to more complex dynamic environments, such as free-fall motion.

In this paper, the temperature-related sensor drift has not been addressed due to the limitations of experimental instruments.
Therefore, future work is planned to maintain consistent calibration accuracy for the gyroscopes across a range of temperatures by incorporating additional weights into TinyGC-Net and taking temperature measurement values as model inputs.
Besides, owing to the powerful modeling capabilities inherent in deep learning models, it is also possible to design a calibration approach for MEMS IMU array modules based on the TinyGC-Net architecture,
which aims to augment the level of automation in the calibration process, thereby enhancing the overall performance of inertial navigation systems.

\section*{Reference}

\bibliographystyle{iopart-num} 

\bibliography{ThisPaperRef.bib}

\end{document}